# Bayesian Neural Networks

| **Vikram Mullachery** | **Aniruddh Khera** | **Amir Husain** |
| --- | --- | --- |
| mv333@nyu.edu | ak5146@nyu.edu | ah3548@nyu.edu |

## Abstract

This paper describes, and discusses Bayesian Neural Network (BNN). The paper showcases a few different applications of them for classification and regression problems.

BNNs are comprised of a Probabilistic Model and a Neural Network. The intent of such a design is to combine the strengths of Neural Networks and Stochastic modeling. Neural Networks exhibit universal continuous function approximator capabilities. Statistical models (also called probabilistic models) allow direct specification of a model with known interaction between parameters to generate data. During the prediction phase, statistical models generate a complete posterior distribution and produce probabilistic guarantees on the predictions. Thus BNNs are a unique combination of neural network and stochastic models with the stochastic model forming the core of this integration. BNNs can then produce probabilistic guarantees on it's predictions and also generate the distribution of the parameters that it has learnt from the observations. That means, in the parameter space, one can deduce the nature and distribution of the neural network's learnt parameters. These two characteristics make them highly attractive to theoreticians as well as practitioners.

Recently there have been a lot of activity in this area, with the advent of numerous probabilistic programming libraries such as: PyMC3, Edward, Stan etc. Further, this area is rapidly gaining ground as a standard machine learning approach for numerous problems.

## Related Work

In addition to early work by C. Bishop[1], and R. Neal[2], there has been recent works by C. Blundell[3], that lead into the recent relevancy of BNN.

## Datasets

As a part of the experimentation here, we will be using three (3) different datasets and related classification and regression problems to train and evaluate a few different flavors of BNN:

| Dataset | Description |
| --- | --- |
| Individual income tax statistics | This data set is based on individual income tax returns provided by IRS. It is aggregated by zip code per agi_stub (which separates the sample sets into 6 based on adjusted gross income).<br><br>https://www.kaggle.com/irs/in |

---

[1] Bayesian Neural Networks [1997] Christopher M. Bishop
[2] Bayesian Training of Backpropagation Networks by the Hybrid Monte Carlo Method [1992] Radford M. Neal
[3] Weight Uncertainty in Neural Networks [2015] Charles Blundell, Julien Cornebise, Koray Kavukcuoglu Daan Wierstra, Google DeepMind

| | dividual-income-tax-statistics |
|---|---|
| US Powerball lottery dataset | Powerball consists of five numbers between 1 and 69 drawn at random without replacement. A sixth "powerball" number is drawn independently of the first five. Various levels of wins are possible, based on the powerball number. The jackpot winning numbers are available here: https://www.kaggle.com/scotth64/powerball-numbers <br><br> The problem setting here is to learn the decision boundary between the first and third numbers in a sorted winning ticket. |
| Technical analysis of S&P 500 companies using Bayesian Regularized Neural Network | In this subsection, a part of kaggle competition https://www.kaggle.com/dgawlik/nyse (technical analysis) is focused to make confident predictions on the next day stock movement. Data for 501 companies' daily stock prices has been provided from 2010 to 2016. |

## Probabilistic Modeling

In probabilistic modeling, one directly specifies a model for the prior parameters of the model and the likelihood, which are then combined to yield the posterior. This modeling assumes a knowledge of the interaction of the parameters in generating observed data. In that sense this is a generative story - and the model parameters combine in a specifiable way to yield the likelihood.

$$\Pr(\theta|y) = \frac{\Pr(y|\theta)\Pr(\theta)}{\Pr(y)}$$

where $\Pr(\theta|y)$ is the Posterior Probability, $\Pr(y|\theta)$ is the Likelihood of Observations, $\Pr(\theta)$ is the Prior Probability, and $\Pr(y)$ is the Normalizing Constant.

Figure 1: Posterior probability computation

The crux of probabilistic modeling is probability conditioning, which is eminently expressed in Bayes Rule shown above. It is assumed that there is a set of unobserved parameters, $\theta$, that define the model. In the figure above, $y$, represents the observed data.

In maximum likelihood modeling, one directly models the $P(y|\theta)$, which is the typical case in a neural network. That is, find the parameters that maximize the probability of the observed data.

However in a probabilistic approach, we use conditioning and assume that parameters ($\theta$) of the model have some distribution according to our prior belief. The parameters, $\theta$, and data, $y$ interact through the likelihood specification. As we observe data ($y$), we compute the posterior parameter distribution as the product of the prior and the likelihood, normalized by the probability of data (this is an intractable quantity, numerically for most interesting cases). The resulting distribution is the posterior distribution of $\theta$ given the observation $y$. Note that this inference gives us a complete probability distribution and we are not dealing with point estimates. Traditionally in probabilistic modeling, the practitioner chooses conjugate prior for the likelihood, so that the posterior can be computed analytically (in a closed form mathematical expression). An example is the Beta prior for a Bernoulli likelihood yields a Beta posterior. Following is a sample listing of conjugate distributions (and parameters) for a few discrete distributions (there exists a similar table for continuous distributions):

**Sample discrete conjugate distributions[4]**

| Likelihood | Prior | Posterior |
|---|---|---|
| Bernoulli p | Beta $\alpha, \beta$ | Beta $\alpha$ - 1 successes, $\beta$ - 1 failures, |
| Categorical **p** (probability vector of k dimensions) | Dirichlet $\alpha$ | Dirichlet $\alpha + \Sigma\ c_i$ $c_i$ is the observations in category i |
| Poisson $\lambda$ | Gamma $\alpha, \beta$ | Gamma $\alpha$ total occurrences in $\beta$ intervals |

However, a conjugacy requirement on the prior becomes a constraint and is not desirable. In such cases we approximate the posterior using sampling or variational inference techniques.

There are three phases to consider in probabilistic modeling - model specification, model inference and model checking.

### Model Specification
This phase includes specifying a prior distribution on the parameters of the model, and the exact nature in which they combine to yield a likelihood function. The model has to be completely defined, such that the priors are not concealed or hidden. Similarly, the likelihood should be a computable function. For the case of a coin toss, where one is uncertain of the coin's fairness - one could establish a distribution on it, that is a degree to which one feels that the coin is fair, and then use this prior parameter to compute the likelihood of the training data. As one sees training data, one's prior beliefs can be updated, based on the computed posterior. If one were to assume that the coin is fair (a uniform or flat prior on the probability of head, say p), and then one sees that a heads show up during a coin toss, we could use Bayes rule to compute the posterior to be 'p' (= p * 1). Simply generalizing this to a situation where one observes 'h' heads and 't' tails, the posterior calculation is as follows:

$$f(r|H=h, T=t) = \frac{\Pr(H=h|r, N=h+t)\, g(r)}{\int_0^1 \Pr(H=h|p, N=h+t)\, g(p)\, dp}.$$

Figure 2: Posterior formulation for coin tosses

where N is the number of coin tosses (h + t). g is the prior probability estimate of the coin tossing a head. If g is assumed to be uniform (g(r) = 1), over the range [0, 1], then, this can be analytically simplified to:

$$f(r|H=h, T=t) = \frac{(h+t+1)!}{h!\, t!}\, r^h\, (1-r)^t.$$

Figure 3: Posterior analytical form for simple coin toss with a uniform prior

The key observation here is that this posterior expression may not be analytically simplifiable for all prior functions, g. In fact in most interesting and realistic scenarios, there will not be a closed form expression for the posterior. Further, computing this expression numerically becomes difficult through any brute force evaluation of all choices (note the presence of the integral in the denominator, which attempts to average out the contribution from all of the parameter space). This compounds in difficulty in the case of multi-dimensional parameter spaces. In most cases, approximation techniques are used: either sample estimates or variational inference techniques.

### Model Inference
The second phase is to compute the posterior value of parameters. This computation is where a lot of historical research and recent advances have been made. For the sake of practitioners, most of the commercial or educational libraries black box this step into a simple easy to use function (or method). There are two families of approaches to consider: sampling methods and variational inference methods.

---

[4] Source: Table of conjugate distributions

## Sampling Methods

Sampling techniques include mostly Markov Chain Monte Carlo family of algorithms, in which the parameter space is sampled in proportion to their probabilities to yield a sample (a collection of parameter values - these could be multidimensional vectors and not simply scalars). Numerous advances have been made in this field since the late 1940s, a few notable ones include: Metropolis Hastings Algorithm, Hamiltonian Monte Carlo, and more recently the No-U-Turn sampler[5] (NUTS).

## Metropolis Hastings Algorithm

Metropolis algorithm is able to sample from the posterior with the knowledge of only the unnormalized measure, **π**, which is the product of the likelihood, p(D|*θ*), and the prior probability, p(*θ*) (for this description D is used to notate data, the same as y from the previous section). This algorithm takes the stance that rather than arbitrarily and randomly sampling points from the parameter space, one could consider a Markov Chain of correlated adjacent states - and then, these states can be sampled so as so yield samples from the desired posterior distribution (without knowing it's normalizing term). Although the Markov Chain is locally correlated, it is possible to make it ergodic (that is the chain will visit every possible state in exactly the proportion that matches its probability mass). This ergodicity requires that we consider a transition probability distribution, p, that will satisfy detailed-balance condition with the unnormalized measure:

**π** ($\theta_0$) p($\theta_1$|$\theta_0$) = **π** ($\theta_1$) p($\theta_0$|$\theta_1$)

---

[5] The No-U-Turn Sampler [2011] Matthew D. Hoffman, Andrew Gelman

Figure 4: Metropolis Algorithm - detailed balance

Detailed balance means that the probability of transition from parameter $\theta_0$ to parameter $\theta_1$ during sampling is the same as going in reverse from $\theta_1$ to $\theta_0$. The proof that ergodicity follows from detailed balance is deceptively simple, however, the concept is quite deep.

All that remains for this algorithm now is to find a transition distribution p. Here the original authors discovered that it is sufficient to find a computationally easy proposal distribution q, (say multivariate Gaussian), and then perform the following algorithmic steps to transition to a next state from current state ($\theta_0$):

- Generate a candidate state $\theta_c$ from q
- Compute an acceptance probability, **α**, such that it obeys the following acceptance/rejection rule:

$$\alpha = \min\left(1, \frac{\pi(\theta_c)q(\theta_0|\theta_c)}{\pi(\theta_0)q(\theta_c|\theta_0)}\right)$$

- Choose the next state with probability **α**, otherwise next state is current state $\theta_0$

The proof that this yields an ergodic sequence follows from the detailed balance property. Simply put, a candidate state is accepted when it is more probable than the current state, otherwise it is accepted part of the time as dictated by the fraction in the parenthesis above

## Hamiltonian Monte Carlo

Hamiltonian Monte Carlo (HMC) method exploits the geometric properties of the typical set (the region with total probability close to 1) of the posterior distribution. Rather than exploring the typical set by means of a random walk (as in Metropolis Hastings algorithm), HMC lifts the exploration problem into a phase space which consists of position (the original parameter, $\theta$) and momentum. The relationship between the position and momentum are described by Hamilton's equations, and can be considered as total energy preserving. HMC has the ability to make far and confident strides in the parameter space, and thus converge faster than Metropolis Hastings method, and also provides stronger guarantees on the resulting estimator[6].

The parameters of HMC are an $\varepsilon$ (step size) and L (number of steps). Samples are generated first by sampling a Standard multivariate Gaussian for momentum parameter (r). Then, L leapfrog updates are made to arrive at a proposal momentum-position parameter pair value ( $\tilde{\theta}, \tilde{r}$ ). Now, we accept or reject this proposed value according to Metropolis algorithm. The probability measure, $\alpha$ used to compute the acceptance is given by:

$$\alpha = \min\left\{1, \frac{\exp\{\mathcal{L}(\tilde{\theta}) - \frac{1}{2}\tilde{r}\cdot\tilde{r}\}}{\exp\{\mathcal{L}(\theta^{m-1}) - \frac{1}{2}r^0\cdot r^0\}}\right\}$$

And if accepted, the new states of position and momentum are set to the proposed values:

$$\text{set } \theta^m \leftarrow \tilde{\theta}, r^m \leftarrow -\tilde{r}.$$

The negative sign in the momentum is required to preserve time-reversibility. The leapfrog-integrator[7] is constructed to be volume-preserving. Together, these two facts ensure that the proposal is a valid Metropolis proposal. The leapfrog-integrator shown below is repeated L (number of steps) times to yield a proposal parameter-pair:

$$\begin{aligned}
&\textbf{function } \text{Leapfrog}(\theta, r, \epsilon) \\
&\text{Set } \tilde{r} \leftarrow r + (\epsilon/2)\nabla_\theta \mathcal{L}(\theta). \\
&\text{Set } \tilde{\theta} \leftarrow \theta + \epsilon\tilde{r}. \\
&\text{Set } \tilde{r} \leftarrow \tilde{r} + (\epsilon/2)\nabla_\theta \mathcal{L}(\tilde{\theta}). \\
&\textbf{return } \tilde{\theta}, \tilde{r}.
\end{aligned}$$

**No-U-Turn Sampler**

NUTS is an adaptive extension of HMC that retains (and sometimes improves on) HMC's efficiency and requires no hand tuning. HMC is a powerful algorithm, but its usefulness is limited by the need to tune the step size parameter, $\varepsilon$ and number of steps, L. Selecting L is particularly problematic, and so practitioners commonly rely on heuristics based on autocorrelation statistics from preliminary runs [2].

The core idea behind NUTS is once the trajectory begins to turn back (towards start), it's time to stop the simulation, putting it formally, stop the simulation when running it longer wouldn't increase the distance (squared) to where we started:

$$\frac{d(.5\,(\theta' - \theta)^2)}{dt} = \left(\frac{d\theta'}{dt}\right)^T (\theta - \theta') = r^T \cdot (\theta' - \theta) < 0$$

where $\theta$ is start position and the $\theta'$ is the current position, rest of the symbols have their usual meaning (from HMC section).

---

[6] A Conceptual understanding to Hamiltonian Monte Carlo, M. Betancourt
[7] Source: The No-U-Turn sampler, M. Hoffman and A. Gelman
[2] Source: MCMC using Hamiltonian dynamics, Radford M. Neal

Pseudo code:
- For each iteration of NUTS proceed by:
    1. $r \sim Normal(0,1)$
    2. (slice variable) $u \sim Uniform([0, p(\theta, r | x)])$
    3. Tracing out Hamiltonian dynamics of $\theta, r$ forwards and backwards in time, going forwards or backwards 1 step, then forwards or backwards 2 steps, then 4 steps, etc
    4. Doubling stops when a sub-trajectory makes a U-turn, at which point we sample from among the points on the trajectory carefully.

The doubling process implicitly builds a balanced binary tree whose leaf nodes correspond to position momentum states, as illustrated in the figure below:

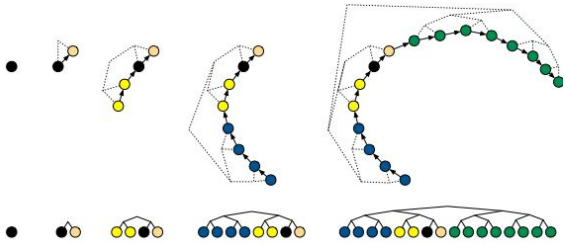

Each doubling proceeds by choosing a direction (forwards or backwards in time) uniformly at random, then simulating Hamiltonian dynamics for $2^j$ leapfrog steps in that direction, where j is the number of previous doublings (and the height of the binary tree). The figures at top show a trajectory in two dimensions (with corresponding binary tree in dashed lines) as it evolves over four doublings, and the figures below show the evolution of the binary tree. In this example, the directions chosen were forward (light orange node), backward (yellow nodes), backward (blue nodes), and forward (green nodes).

Until now we talked about selecting the L. To chose ε for both NUTS and HMC, stochastic optimization with vanishing adaptation, specifically an adaptation of the primal-dual algorithm of Nesterov[10] has been proposed.

## Variational Inference Methods

The underlying idea in variational inference methods is to translate the inference of the posterior into an optimization (minimization or maximization) problem. A variational distribution, q(**θ**; **ν**), parameterized by **ν** is introduced which attempts to closely approximate the posterior, p(**θ**|x). The variational family from which, q, is chosen needs to be flexible enough to represent a wide variety of distributions and able to capture the posterior, p. Kullback-Leibler (KL) divergence between p and q, over q, is used to measure this closeness. So applying the definition of KL divergence to the posterior we arrive at:

$$KL(q||p) = \mathbb{E}_q[\log \frac{q(\theta;\nu)}{p(\theta|x)}]$$

Figure 4: KL divergence between the variational distribution and the posterior

which can be further simplified to:

$$\underbrace{KL(q||p)}_{\text{KL divergence}} = -(\underbrace{\mathbb{E}_q[\log p(\theta, x)]}_{\text{exp. log joint}} - \underbrace{\mathbb{E}_q[\log q]}_{\text{entropy}}) + \underbrace{\log p(x)}_{\text{Constant}}$$
$$\text{ELBO, } \mathscr{L}$$

Figure 5: Evidence Lower bound

Minimizing the KL divergence is now equivalent to maximizing the evidence lower bound (ELBO), since they sum up to an unknown but constant value, as in the equation above. This is so because the probability of data (x) is the intractable but constant value, p(x).

There are two forces at play here in the maximization of ELBO, the expectation of the log joint distribution of **θ**, x under q and the entropy of q. The first term drives the variational distribution to place weights where the posterior distribution is high. This first term is similar to a maximum a posteriori estimate (MAP) of the posterior. On the other hand, the second term attempts to make the variational distribution diffuse and place weights widely around it's parameter space, **ν**. Since the intent is to maximize the ELBO, one could use optimization techniques based on gradient ascent. And so, it is important to realize that the ELBO maximization objective is non convex, and does not involve computing a global optimum.

Pictorially, variational inference can be seen as a search for the optimal parameter **ν*** of the variational distribution that minimizes the distance (KL divergence between q and p), as in figure below:

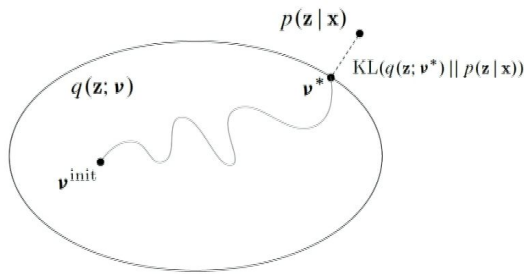

Minimize KL between $q(\beta, z; \nu)$ and the posterior $p(\beta, z|x)$.

**Figure 6: Maximization of ELBO pictorially[8]**

Stochastic update techniques have been applied to variational inference that allow it to scale to large datasets[9]. These stochastic updates work using noisy (but asymptotically unbiased) estimators of the gradient at each iteration by considering a single (or a mini-batch of) data point(s). This is akin to the familiar and all too popular Stochastic Gradient Descent (SGD) used in optimizing the objective functions (also called loss or cost functions) of neural networks.

Mean field variational inference makes simplifying assumptions to the model that the components of the parameters can be fully factorized to independent parts and thus allow for simpler modeling regime. This boils down to, mathematically, that the variational distribution q(z) = **Π** $z_i$, and where the $z_i$'s are considered independent.

Despite its expressive power, variational inference still requires model specific derivatives and implementations. And that's where automatic differentiation variational inference comes into play.

**Automatic Differentiation Variational Inference (ADVI)**

ADVI builds automated solutions to variational inference by transforming the space of latent variables and by automating derivatives of the joint distribution. So, the inputs are the probability model and the dataset, and the outputs are the posterior inferences about the model's latent variables.

The exact steps involved in ADVI are[10]:

1. Transform the model from p(x, **θ**) to unconstrained real number valued random variables, p(x, ζ). Now the variational optimization problem is defined on the transformed problem that is to minimize KL (q(ζ) ∥ p(ζ|x)). All latent variables are now defined in the same real space and ADVI can now use a single variational family for all probabilistic models.
2. Recast the gradient of the variational objective (which includes the gradient of the log joint as shown below) as an expectation over q. Once this gradient is expressed as an expectation, it allows us to use Monte Carlo

---

[8] Courtesy D. Blei, Columbia University
[9] Stochastic Variational Inference, M. D. Hoffman, D. M. Blei, C. Wang, J. Paisley
[10] Stochastic primal-dual, Anatoli Juditsky and Yuri Nesterov

[10] Automatic Differentiation Variational Inference, A. Kucukelbir, D. Tran, R. Ranganath, A. Gelman, D. Blei

methods to approximate it:

$$\nabla_\theta \log p(x, \theta)$$

3. ADVI further re-parameterizes the gradient in terms of a Gaussian by transforming once again within the variational family. This transformation allows for efficient Monte Carlo approximations
4. ADVI uses stochastic optimization for the variational distribution

A few points worth noting about ADVI are that there are certain models that it cannot be applied to, because of the intractability of marginalizing the discrete variables in the likelihoods of such models. These include: Ising model, sigmoid belief network, and Bayesian nonparametric models. These non-differentiable probability models are not candidates for ADVI. The mathematical formulation for the objective of ADVI is illustrated in the Proofs and Formulas section.

**Model Checking**

The third phase of probabilistic modeling is model checking and comparison, which includes checking sensitivity to prior distributions.

Asymptotic guarantees of posterior distribution convergence independent of the prior is quite advantageous, under conditions of the random variables having a finite probability space. These guarantees also require an independent and identically distributed sample space assumption[11].

Nevertheless, it is important to perform experimentation with various prior distributions, especially uninformative priors, such as uniform or flat priors, and at times even improper priors. Typically, the posteriors should be robust to changes in the prior. If that is not the case, then it is highly likely that the model itself may not be apt or correct, and should be reason to explore alternatives.

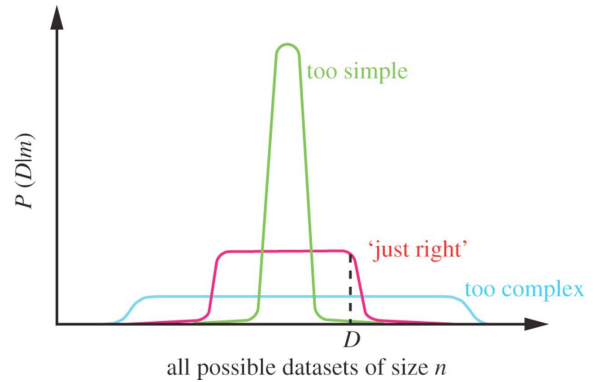

Figure 6: Occam's razor at work

In the above figure we plot the likelihood of data on the y-axis against the spread of the data in input space for a given dataset size, n (a simplified toy example). What one can notice is that a complex model (blue curve) will have very little probability mass for P(D|m) (the likelihood), because the model is capable of representing far more spread out data. Contrarily, a simple model (green curve) can only span a small region of the input space (horizontal axis), and thus again can only apportion very little probability measure to the data. When the model is of the correct complexity (purple curve) and can span the dataset, it is able to apportion a large part of it's probability mass to likelihood. This is a natural expression of Occam's razor principle and falls in line with probabilistic modeling, because all models have to work with the same amount of probability measure (that is a total of 1).

In conventional machine learning computational models this notion of goodness of fit is bolstered by a regularization mechanism. In Deep Learning models, regularization is obtained using weight decay parameter and dropouts, among other techniques (sparsity constraints, model parameter constraints etc.). Traditional machine learning models further utilize held out data sets, such as validation data sets as a check against overfitting.

---

[11] Bernstein-von Mises Theorem

$$P(m|\mathcal{D}) = \frac{P(\mathcal{D}|m)P(m)}{P(\mathcal{D})}$$

$$P(\mathcal{D}|m) = \int P(\mathcal{D}|\theta, m)P(\theta|m)\,d\theta$$

**Figure 5: Integrating over all parameters θ, yields the likelihood of the data for a given model, m**

With probabilistic modeling, there is no necessity to inject a regularizer or even a validation dataset, albeit it is done in cases where a side by side comparison with other models may be desired.

Notwithstanding all that, model checking is absolutely essential. Model evidence P(D|m) can be used for this, since models that have higher marginal data likelihoods (also called model evidence) are able to account for data more adequately. The ratio of the model evidences of two models is called the Bayes factor and can be used to compare models.

An additional check that practitioners use is generating samples from the posterior distribution, and comparing them to the original data. This is done using posterior predictive check samples, and their closeness to the original data. Usually metrics such as highest posterior density and posterior predictive map are compared to the variance and mean of the sample.

## Neural Networks

Neural networks are simple to understand. They consist of an artificial neuron, which performs a linear transformation of the input values and a nonlinearity (typically a sigmoid function).

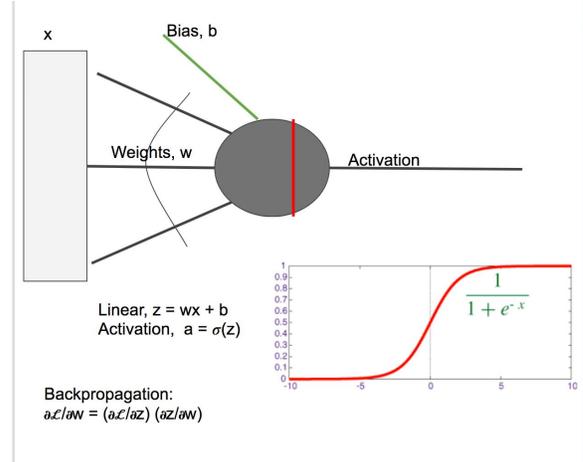

**Figure 7: Artificial Neuron**

Note that there are weights associated with each of the inputs to a neuron. As shown in the figure, the input values are weighted by the associated weight and then summed together. This summed scalar value is passed through the nonlinearity to arrive at the output scalar value of a neuron.

When numerous artificial neurons are stacked together to form a layer and many such layers are placed adjacent to each other we get a simple neural network as shown below (simplified):

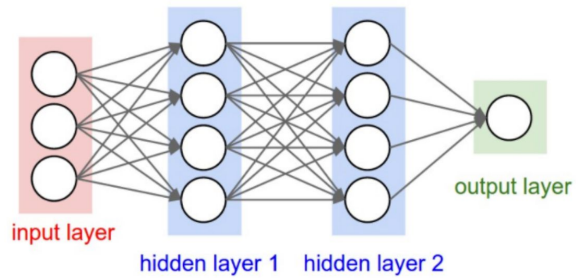

**Figure 8: Neural Network showing hidden layers and a single output[12]**

---

[12] Courtesy Andrej Karpathy at Stanford University

In a supervised learning setting, the input vectors (of size 3 in the figure above), are supplied to the network on the left. The network weights are let's say randomly initialized. The network computes the output of each of the neurons at the first layer and then propagates the results to the second layer, where a similar computation is performed and so on, until the last layer. At the last layer (in figure we show only a single scalar output, however, one could construct a vector of multiple scalar values as outputs), the network predicts a particular value. In a supervised setting we know the label or the expected outcome. A loss function computes the deviation of the predicted network outcome (ŷ) from the expected outcome (y). This loss value is back propagated, through the layers of the network so that at each layer the network adjusts its weights to the extent that it contributed to the error in the output. Mathematically, this is an application of chain rule of partial differentiation. There is an expectation of differentiability (mostly) in order for this technique to work. An optimization algorithm such as Stochastic gradient descent (SGD) is used to compute the loss and back propagate it to the weights in the neural network. This activity is performed repeatedly (in batches mostly) to minimize the loss at each successive round. Termination criteria is usually the lowest observed cost (or error) on a separate cross-validation data set, against which the network is not trained but evaluated. If the loss on the cross validation dataset tends to increase then one stops training the network any further. This learnt model is now ready for prediction on a test dataset.

As one can easily imagine there is a wealth of detail that we have glossed over to simplify the discussion.

When posed with new input (data/vectors) from a test set, the neural net computes the output and presents it as the prediction.

This outcome prediction has no warranties about the network's sense of certainty or variability of predicted outcome, it is a point prediction. Without further engineering, one is typically not able to make claims about how much the output would vary by varying a particular input vector component, say the 3rd component (or feature). If the input vector elements can be considered as features, then questions of feature relevance (to the output) and inter-feature correlation might be important. However, without additional scaffolding, such inferences are not easily or directly obtainable.

Further and more importantly, the neural network has no "deducible" model of the world that it has learnt that can be used by the practitioner. Yes there are weight vectors and their network arrangement, however there is no direct correlation to any statistical model. The distribution of individual weight parameters (across the neural network layers), in the parameter space is unknown; all we know are its point values at the termination of training.

Nevertheless, neural networks have proven to be very good at image recognition and computer vision tasks. The sheer size of their parameter space, in the order of hundreds of millions, easily tell us that they are endowed with a very large parameter space, with capabilities to learn highly complex relationships between input and output.

## Bayesian Neural Networks

In order to combine the best of neural networks and probabilistic modeling, researchers innovated BNN. The goal is to benefit from the bounds and guarantees of probabilistic modeling and ability for neural networks to be universal function approximators.

In BNNs usually, a prior is used to describe the key parameters, which are then utilized as input to a neural network. The neural networks' output is utilized to compute the likelihood, again with a specific defined probability distribution. From this, one computes the posterior distribution of the parameters by sampling or variational inference.

A BNN is at its core a probabilistic model augmented with neural network as a universal function approximator. It is the very nature of statistical modeling to generate distributional outputs and this is truly what allows a BNN to produce more than a point estimate.

## Experiments

We attempted three different types of problems using Bayesian Neural Networks. We used the python library PyMC3[13] for our experimentation. Following is a description of the experimentation. The code for our experimentation is available at our code repository[14] on github.

### Powerball dataset

The US powerball dataset consists of a list of the winning powerball ticket numbers. There are six lottery ticket numbers, the first five numbers are drawn at random without replacement from 1 to 69 (and are white colored balls - in the actual lottery drawing). The sixth number (a red colored ball), also called powerball number, is drawn independently at random from a pool of 26 numbers[15]. The draws are twice a week on Wednesdays and Saturdays evenings. If a ticket matches all the 6 numbers (the first five white balls, and the sixth red ball), it is a jackpot.

Following is a raw data plot of powerball numbers in positions 1 and 3:

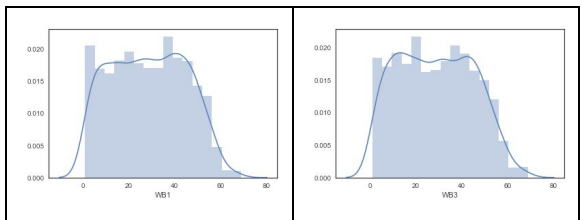

**Figure 10: Unsorted winning numbers in positions (bins) 1 and 3**

Since the first five numbers are picked from the same range without replacement, one could sort them and study their distribution. It is immediately reminiscent of a Poisson-like distribution for bins 1 and 3 (and so for the rest as well, as below):

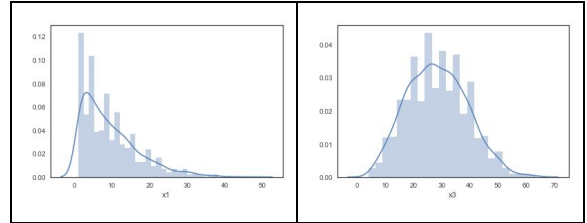

**Figure 11: Frequency of winning numbers in the bins 1 and 3. Notice the overlap region on the x-axis**

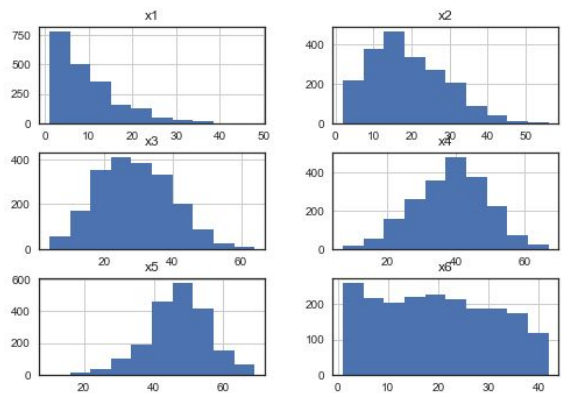

**Figure 12: Frequency of winning numbers in each of the six sorted buckets**

For this task, we wish to predict the probability of a number being picked from bin 1 or bin 3. So, given a powerball winning lottery ticket number (one amongst the first five whiteball numbers), whether it will be possible to predict whether number belongs to bin 1 or bin 3? This is a classification problem. We were additionally interested in the certainty that we would assign to our predictions.

We compared our BNN for accuracy of prediction against a Random Forest Classifier, a Gaussian Classifier and a AdaBoost decision tree.

### Accuracy on held-out data

| Model Description | Accuracy |
| --- | --- |

---

[13] PyMC3 - Probabilistic modeling in Python
[14] MLExp - Bayesian Neural Network
[15] US Powerball

| | |
|---|---|
| BNN with 1 hidden layer, 5 units each, tanh activation and log-transformed input | 81.23% |
| Random Forest<br>10 estimators, max depth 5 | 82.96% |
| AdaBoost Decision Trees | 82.96% |
| Gaussian Process Classifier (Radial Basis Function kernel) | 84.93% |

## Classification Boundary

The classification surface learnt by these different models were various and allow us to perceive some of their functional differences. The smoothness in the functional form of these classifiers are evident from the boundary plots:

| Model | Boundary |
|---|---|
| BNN | |
| Random Forest | |
| AdaBoost Decision Trees | |
| Gaussian Process Classifier (Radial Basis Function kernel) | 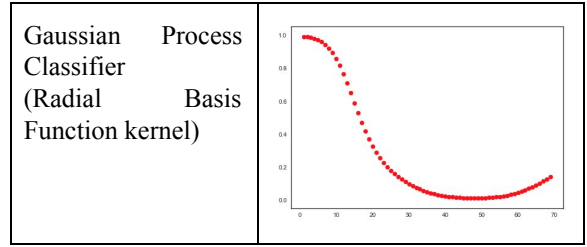 |

## Prediction Uncertainty

Bayesian Neural Networks automatically yield us the prediction certainties. From the below plot it is obvious that the prediction uncertainty is low for the ball being from the first bin (green error bar) when the number is between 1 and 10. We notice the uncertainty increase as we move from 1 to 10, and at around 14, the uncertainty of being from the first bin is matched by the uncertainty of the ball originating from the third bin (green bar + red bar). Reading farther right in the plot we see that the uncertainty of the ball being from the first bin reduces (and the error bars turn red).

A curious observation is the high uncertainty for the number zero. If one pays close attention, one would notice that zero is not a valid number within our training set and is in fact never observed. However, when we force the classifier to make a choice for zero (due to sampling), it comes back with an extremely high uncertainty for both being in bin 1 and in bin 3. This further validates and affirms our faith in the model predictions.

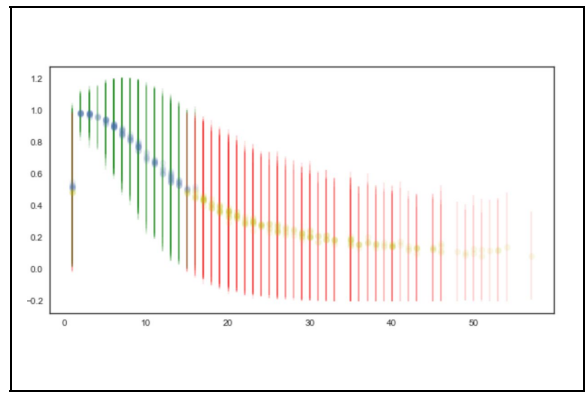

## Confusion Matrix

False positives in predicting the first bin (= 1, here) is 26% is higher than the false negatives (=0, here) around 6.8%. This tells us that the model is quite aggressive in claiming that a ball came from the first bin than it should be.

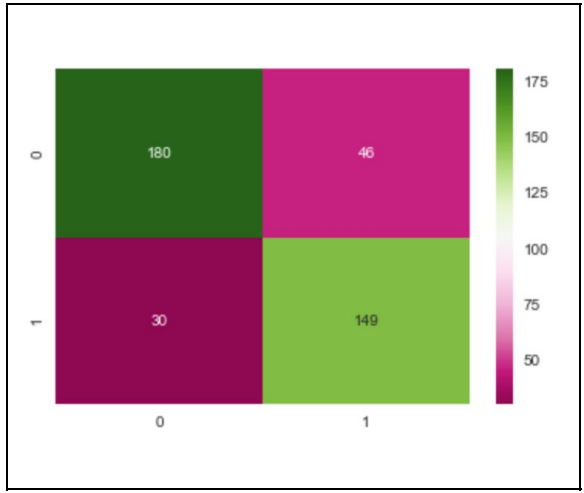

## Learnt neural network weights

It is highly instructive to visualize the distribution of the neural net weights in the three layers: input, hidden and output:

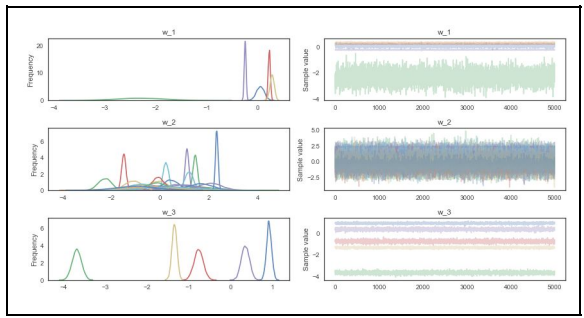

Some of the weights exhibit very narrow bases and they can be interpreted as having learnt a weight that has a narrow highest posterior density. This tells us that that particular neuron weight is decidedly certain, and less sensitive to the input variations that it has observed. On the other hand, some weight posteriors have wide bases and are highly sensitive to the input data that it has encountered, and thus exhibit high variance.

Such insights are priceless for a deep learning practitioner, should they choose to interpret the meanings of the weight vectors and the input features. These insights are unique to BNNs.

### S&P 500 Dataset

Currently the work on this dataset is not publishable. We are exploring available libraries that allow us to construct a bayesian neural network with autoregressive models.

### Income tax Dataset

Currently the work on this dataset is not complete.

## Conclusion

Bayesian Neural Network (or Bayesian Deep Learning) exposes a few powerful techniques and insights into deep learning. It enhances regular neural networks with predictive uncertainties and posterior network weight distributions, which are both handy in feature learning and in model building.

The backbone of BNN is probabilistic modeling. The key ingredient that combines neural networks with probabilistic modeling is the posterior approximation technique. Sampling is a traditional approximation technique. A contemporary method that has seen great activity is variational inference (VI) based methods such as: Stochastic VI, Blackbox VI, ADVI (discussed in this paper), operator variational inference (OPVI). These areas have seen great resurgence in research and innovation, and offer great promises. BNNs have good principled approaches to support the modeling and the exhibited results, particularly in comparison to vanilla deep learning models.

Though BNNs combine the powerful features of probabilistic modeling and neural networks, they also suffer from some of their weaknesses. Currently, BNNs are computationally expensive, because of the sampling or variational inference steps. BNNs have been demonstrated to be competent on moderately sized datasets and not yet fully explored with vastly large datasets.

## Future work

Some of the areas of active research that directly impact BNNs:

- Exploration of alternate divergences (than KL divergence)
- Exploration of variational inference techniques
- Research in areas of approximations
- Work in the area of non-convexity of evidence lower bound (variational inference)

## Proofs and Formulas

### Detailed Balance and Ergodicity

**Theorem**: *If a proposal probability distribution q, satisfies detailed-balance condition with the unnormalized probability measure, π, then it necessarily means that the Markov Chain is ergodic with probability measure π*

**Proof**: Detailed balance implies:

$\pi(\theta_0) q(\theta_1|\theta_0) = \pi(\theta_1) q(\theta_0|\theta_1)$

Now, computing the probability of being in a successor state (or parameter) of $\theta_0$ is proportionally given by integrating over all $\theta_0$, the product of probability of being in $\theta_0$ and transitioning to $\theta_1$:

$\int \pi(\theta_0) q(\theta_1|\theta_0) d\theta_0$

By detailed balance, this simplifies to:

$\pi(\theta_1) \int q(\theta_0|\theta_1) d\theta_0 = \pi(\theta_1)$

[because q is a proper probability distribution].

That proves the theorem.

### Metropolis Hastings Algorithm and Ergodicity

**Theorem**: *The following algorithmic steps in Metropolis sampling yields an ergodic sequence*

1. Generate a candidate point $x_{2c}$ by drawing from the proposal distribution around $x_1$
2. Calculate an "acceptance probability" by

$$\alpha(x_1, x_{2c}) = \min\left(1, \frac{\pi(x_{2c}) q(x_1|x_{2c})}{\pi(x_1) q(x_{2c}|x_1)}\right)$$

Notice that the q's cancel out if symmetric on arguments, as is a multivariate Gaussian

3. Choose $x_2 = x_{2c}$ with probability $\alpha$, $x_2 = x_1$ with probability $(1-\alpha)$

So, $p(x_2|x_1) = q(x_2|x_1) \alpha(x_1, x_2)$, $(x_2 \neq x_1)$

**Proof**: [Courtesy Prof. William H. Press]

$\alpha$ be the acceptance probability for transition from state $x_1$ to a candidate state $x_{2c}$

Proof:
$$\alpha(x_1, x_{2c}) = \min\left(1, \frac{\pi(x_{2c}) q(x_1|x_{2c})}{\pi(x_1) q(x_{2c}|x_1)}\right)$$

So,
$$\pi(x_1) q(x_2|x_1) \alpha(x_1, x_2) = \min[\pi(x_1) q(x_2|x_1), \pi(x_2) q(x_1|x_2)]$$
$$= \min[\pi(x_2) q(x_1|x_2), \pi(x_1) q(x_2|x_1)]$$
$$= \pi(x_2) q(x_1|x_2) \alpha(x_2, x_1)$$

But
$p(x_2|x_1) = q(x_2|x_1) \alpha(x_1, x_2)$, $(x_2 \neq x_1)$
and also the other way around

So,
$$\pi(x_1) p(x_2|x_1) = \pi(x_2) p(x_1|x_2)$$
which is just detailed balance, q.e.d.

Professor William H. Press, Department of Computer Science, the University of Texas at Austin

The steps follow from:

- Multiplying both sides of the equation by the positive denominator
- switching the sequence of variables in a min function
- symmetry of the equation with respect to $x_1$ and $x_{2c}$

### ADVI objective function

**Initial ELBO**

The initial objective of maximizing the evidence lower bound (ELBO) is:

$$\mathcal{L}(\phi) = \mathbb{E}_{q(\theta)}[\log p(x, \theta)] - \mathbb{E}_{q(\theta)}[\log q(\theta; \phi)].$$

where p is the joint and q, the variational distribution. The variational distribution is parameterized by $\phi$, and the original model parameters are $\theta$.

**Variational Problem in Real Coordinate Space**

Following the first step in ADVI, when $\theta$ is transformed by T to real space, we get:

$$\zeta = T(\theta).$$

And so now the joint can be described as:

$$p(x, \theta = T^{-1}(\zeta))$$

, and the transformed joint density is:

$$p(x, \zeta) = p(x, T^{-1}(\zeta)) |\det J_{T^{-1}}(\zeta)|,$$

where $J_{T^{-1}}(\zeta)$ is the Jacobian of the inverse of T. In this transformed space, we use a Gaussian distribution for variational approximation. One could consider a mean-field Gaussian variational approximation in k real dimensions of $\zeta$. Another option is to posit a full-rank Gaussian variational approximation:

$$q(\zeta; \phi) = \mathcal{N}(\zeta; \mu, \Sigma)$$

To ensure that $\Sigma$ always remains, positive semidefinite, we re-parameterize the covariance matrix using a Cholesky factorization $\Sigma = LL^T$

If we use the non-unique definition of Cholesky, L lives in the unconstrained space of lower-triangular matrices with K(K+1)/2 real values and allows negative diagonal elements.

$$q(\zeta; \phi) = \mathcal{N}(\zeta; \mu, LL^T)$$

Now the variational objective in the real coordinate space is:

$$\mathcal{L}(\phi) = \mathbb{E}_{q(\zeta;\phi)}\left[\log p(x, T^{-1}(\zeta)) + \log|\det J_{T^{-1}}(\zeta)|\right] + \mathbb{H}[q(\zeta; \phi)].$$

**Elliptical Standardization**

Consider a transformation that absorbs the variational parameters $\phi$:

$$\eta = S_\phi(\zeta) = L^{-1}(\zeta - \mu).$$

The variational objective now becomes:

$$\phi^* = \arg\max_\phi \mathbb{E}_{\mathcal{N}(\eta;0,I)}\left[\log p\left(x, T^{-1}(S_\phi^{-1}(\eta))\right) + \log|\det J_{T^{-1}}(S_\phi^{-1}(\eta))|\right] + \mathbb{H}[q(\zeta; \phi)]$$

The expectation is now over a standard Gaussian. The Jacobian of the elliptical standardization evaluates to one.

**Stochastic Optimization**

Since the expectation in the above expression does not depend on $\phi$, we can push the gradient through the expectation to yield:

$$\nabla_\mu \mathcal{L} = \mathbb{E}_{\mathcal{N}(\eta)} \left[ \nabla_\theta \log p(x, \theta) \nabla_\zeta T^{-1}(\zeta) + \nabla_\zeta \log \left| \det J_{T^{-1}}(\zeta) \right| \right]$$

and

$$\nabla_L \mathcal{L} = \mathbb{E}_{\mathcal{N}(\eta)} \left[ \left( \nabla_\theta \log p(x, \theta) \nabla_\zeta T^{-1}(\zeta) + \nabla_\zeta \log \left| \det J_{T^{-1}}(\zeta) \right| \right) \eta^\top \right] + (L^{-1})^\top$$

Now, one can compute the gradient inside the expectation using automatic differentiation. In order to calculate the expectation, one can use MC integration; drawing samples from a standard Gaussian, and evaluating empirical mean of the gradients within the expectation. The above gives noisy but unbiased estimators of the ELBO gradient for any differentiable probability model. One can now use these gradients within a stochastic optimization routine to automate variational inference.

## Acknowledgement

## References

1. Weight Uncertainty in Neural Networks [2015] Charles Blundell, Julien Cornebise, Koray Kavukcuoglu Daan Wierstra, Google DeepMind
2. MCMC using Hamiltonian dynamics [2012] Radford M. Neal
3. Bayesian Neural Networks [1997] Christopher M. Bishop
4. Bayesian Training of Backpropagation Networks by the Hybrid Monte Carlo Method [1992] Radford M. Neal
5. The No-U-Turn Sampler: Adaptively Setting Path Lengths in Hamiltonian Monte Carlo [2011] Matthew D. Hoffman, Andrew Gelman
6. Markov Chain Monte Carlo and Variational Inference: Bridging the Gap [2015] Tim Salimans, Diederik P. Kingma and Max Welling
7. Dropout Inference in Bayesian Neural Networks with Alpha-divergences [2017] Yingzhen Li Yarin Gal
8. Automatic Differentiation Variational Inference, [2016] A. Kucukelbir, D. Tran, R. Ranganath, A. Gelman, D. Ble
9. Stochastic optimization, Robbins and Monro, 1951